\documentclass[a4paper,twoside]{article}

\usepackage{epsfig}
\usepackage{subcaption}
\usepackage{calc}
\usepackage{amssymb}
\usepackage{amstext}
\usepackage{amsmath}
\usepackage{amsthm}
\usepackage{multicol}
\usepackage{pslatex}
\usepackage{apalike}

\let\bibhang\relax
\usepackage{natbib}
\usepackage{physics} 
\usepackage[detect-all,locale=DE]{siunitx} 
\usepackage[nice]{nicefrac}
\usepackage[table,dvipsnames]{xcolor} 
\usepackage{color}
\usepackage{tikz}
\usetikzlibrary{%
				automata,%
				arrows,%
				backgrounds,%
				calc,%
				chains,%
				decorations,%
				decorations.markings,%
				decorations.pathmorphing,%
				external,%
				fadings,%
				fit,%
				matrix,%
				positioning,%
				scopes,%
				shadows,%
				shapes.geometric,%
				shapes.misc,%
				shapes.multipart,%
				through,%
				mindmap, 
				backgrounds, 
				topaths}
\usepackage{tikz-3dplot}

\usepackage{multirow}
\usepackage[]{graphicx}
\usepackage{transparent}

\usepackage{epsfig}
\usepackage{epstopdf}
\usepackage[]{todonotes}
\usepackage{booktabs}									

\usepackage{sidecap}

\usepackage[ruled]{algorithm2e}
\usepackage{anyfontsize}

\usepackage[backref=page,pagebackref=true,breaklinks=true,colorlinks,bookmarks=false, linkcolor=black, citecolor=Fraunhofergreen, urlcolor  = Fraunhofergreen]{hyperref}

\definecolor{KITred}{RGB}{160,30,40}

\definecolor{KITblue}  {RGB}{ 70,100,170} 
\definecolor{KITblue70}{RGB}{125,146,195} 
\definecolor{KITblue50}{RGB}{162,177,212} 
\definecolor{KITblue30}{RGB}{199,208,229} 
\definecolor{KITblue15}{RGB}{227,232,242} 

\definecolor{KITlilac}{RGB}{160,0,120}
\definecolor{immCV}{RGB}{170,166,57}
\definecolor{immCA}{RGB}{60,49,118}
\definecolor{Fraunhofergreen}{RGB}{23,156,125}
\definecolor{Fraunhoferblue}{RGB}{31,130,192} 
\definecolor{Fraunhofersteelblue}{RGB}{0,91,127}
\definecolor{Fraunhofersilvergrey}{RGB}{166,187,200}  
\definecolor{Fraunhoferorange}{RGB}{245,130,32}  
\definecolor{Fraunhofergraphit}{RGB}{28,63,82}  
\definecolor{Fraunhoferred}{RGB}{226,0,26} 
\definecolor{Fraunhofer_sand}{RGB}{211,199,174} 
\definecolor{Fraunhofer_weinrot}{RGB}{124,21,77} 
\definecolor{colormaplow}{RGB}{253,231,37} 
\definecolor{colormaphigh}{RGB}{68,13,84}

\usepackage{SCITEPRESS}     
\begin{document}

\title{Generating Synthetic Training Data for \\ Deep Learning-Based UAV Trajectory Prediction.}


\author{\authorname{Stefan Becker \sup{1}\orcidAuthor{0000-0001-7367-2519}\href{mailto:stefan.becker@iosb.fraunhofer.de}{\includegraphics[scale=0.04]{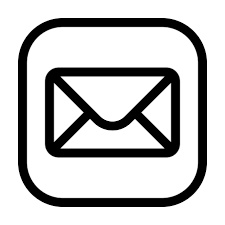}}, Ronny Hug\sup{1}\orcidAuthor{0000-0001-6104-710X}, Wolfgang Huebner \sup{1}\orcidAuthor{0000-0001-5634-6324}, \\ Michael Arens \sup{1}\orcidAuthor{0000-0002-7857-0332}, 
Brendan T. Morris \sup{2} \orcidAuthor{0000-0002-8592-8806}}
\affiliation{\sup{1}Fraunhofer IOSB, Ettlingen, Germany\\Fraunhofer Center for Machine Learning}
\affiliation{\sup{2}University of Nevada, Las Vegas, USA}
\email{\{firstname.lastname\}@iosb.fraunhofer.de, brendan.morris@unlv.edu}
}


\abstract{Deep learning-based models, such as \emph{recurrent neural networks} (RNNs), have been applied to various sequence learning tasks with great success. Following this, these models are increasingly replacing classic approaches in object tracking applications for motion prediction. On the one hand, these models can capture complex object dynamics with less modeling required, but on the other hand, they depend on a large amount of training data for parameter tuning. Towards this end, we present an approach for generating synthetic trajectory data of \emph{unmanned-aerial-vehicles} (UAVs) in image space. Since UAVs, or rather quadrotors are dynamical systems, they can not follow arbitrary trajectories. With the prerequisite that UAV trajectories fulfill a smoothness criterion corresponding to a minimal change of higher-order motion, methods for planning aggressive quadrotors flights can be utilized to generate optimal trajectories through a sequence of $3$D waypoints. By projecting these maneuver trajectories, which are suitable for controlling quadrotors, to image space, a versatile trajectory data set is realized. To demonstrate the applicability of the synthetic trajectory data, we show that an RNN-based prediction model solely trained on the generated data can outperform classic reference models on a real-world UAV tracking dataset. The evaluation is done on the publicly available \emph{ANTI-UAV} dataset.}

\onecolumn \maketitle \normalsize \setcounter{footnote}{0} \vfill
\section{\uppercase{Introduction}}
\label{sec:intro} 
The rise of \emph{unmanned-aerial-vehicles} (UAVs), such as quadrotors, in the consumer market has led to concerns about associated potential risks for security or privacy. The potential intended or unintended misuses pertain to various areas of public life, including locations, such as airports, mass events, or public demonstrations \citep{Laurenzis_SPIE_2020}. Thus, automated UAV detection and tracking have become increasingly important for security services for anticipating UAV behavior. Video-based solutions offer the benefit of covering large areas and are cost-effective to acquire \citep{Sommer_AVSS_2017}. The basic components of such a video-based approach are the appearance model and the prediction model which is traditionally realized with Bayesian filter. The prediction model tasks within detection and tracking pipelines, among others, are the prediction of the object behavior and bridging detection failures. 
Following the success of deep learning-based models in various sequence processing tasks, these models become the standard choice for object motion prediction. A significant drawback of deep learning models is the requirement of a large amount of training data.\\  
In order to overcome the problem of limited training data in the context of UAV tracking in image sequences, this paper proposes to utilize methods from planning aggressive UAV flights to generate suitable and versatile trajectories. The synthetically generated $3$D trajectories are mapped into image space before they serve as training data for deep learning prediction models.\\
Despite the increasing number of trajectory datasets for object classes like pedestrians (e.g., \emph{TrajNet++} dataset \citep{Kothari_arXiv_2020}) or vehicles (e.g., \emph{InD} dataset \citep{Bock_arXiv_2019}), datasets with UAV trajectories and UAV in general are very limited. For the aforementioned object classes of pedestrians and vehicles, mostly RNN-based deep learning models are successfully applied for trajectory prediction \citep{Alahi_CVPR_2016}. Independent of the existing deep learning variants for trajectory prediction relying on \emph{generative adversarial networks} (GANs) \citep{Amirian_CVPR_W_2019,Gupta_CVPR_2018}, \emph{temporal convolution networks} (TCNs) \citep{Becker_ECCVW_2018,Nikhil_ECCVW_2018}, and \emph{transformers} \citep{Giuliari_ICPR_2020,Saleh_arXiv_2020}, an RNN-based prediction model is chosen as reference. The reader is referred to these surveys \citep{Rasouli_arXiv_2020,Rudenko_IJRR_2020,Kothari_arXiv_2020} for a comprehensive overview of current deep learning-based approaches for trajectory prediction.\\
Although the focus is on motion prediction and not on appearance modeling for UAV detection and tracking, we throw a brief glance at some of the current approaches that rely on images or use other modalities. Besides image-based methods, common modalities for UAV detections are RADAR, acoustics, radio-frequencies, and LIDAR. Comparisons of key characteristics of different deep learning-based approaches on single or fused modality information are presented in \citet{Stamatios_MPDI_2019,Taha_IEEEAccess_2019,Unlu_TCVA_2019}. A review with the focus on radar-based UAV detection methods is given by \citet{Christnacher_SPIE_2016}. Approaches for acoustic sensors are, for example, presented in the work of \cite{Kartashov_TCSET_2020} and \cite{Jeon_EUSIPCO_2017}. An example to detect and identify UAVs based on their radio frequency signature is the approach of \cite{Xiao_ICSAI_2019}. For LIDAR, there exist, for example, the approaches of \cite{Hammer_SPIE_2019,Hammer_SPIE_2018}. Image-based approaches can be divided into using electro-optical sensors or infrared sensors. Their appearance modeling, however, is very similar. A further division can be made into one-stage strategies or two-stage strategies. In one-stage strategies, a direct classification and localization is applied. In two-stage strategies, a general (moving) object detection is followed by a classification step. For fixed cameras, the latter strategy is preferred. Different image-based approaches are, for example, presented in \cite{Schumann_SPIE_2018,Schumann_AVSS_2017}, \cite{Sommer_AVSS_2017}, \cite{Mueller_SPIE_2019}, and \cite{Rozantsev_TPAMI_2017}.\\
The paper is structured as follows. Section \ref{sec:data_gen} presents the proposed methods for generating realistic UAV trajectory data in image sequences. Section \ref{sec:model} briefly introduces the used RNN-based UAV trajectory prediction model. In addition to an analysis of the diversity of the synthetically generated data, section \ref{sec:eval} includes an evaluation on the real-world \emph{ANTI-UAV} dataset \citep{Jiang_arXiv_2021}.  Finally, a conclusion is given in section \ref{sec:conclusion}.

\begin{figure*}[!ht]
\centering
\resizebox{1.0\textwidth}{!}{
				\begin{tikzpicture}[]
					\begin{scope}							
					\node (image1) at (-4,0) {	\includegraphics[width=0.5\textwidth]{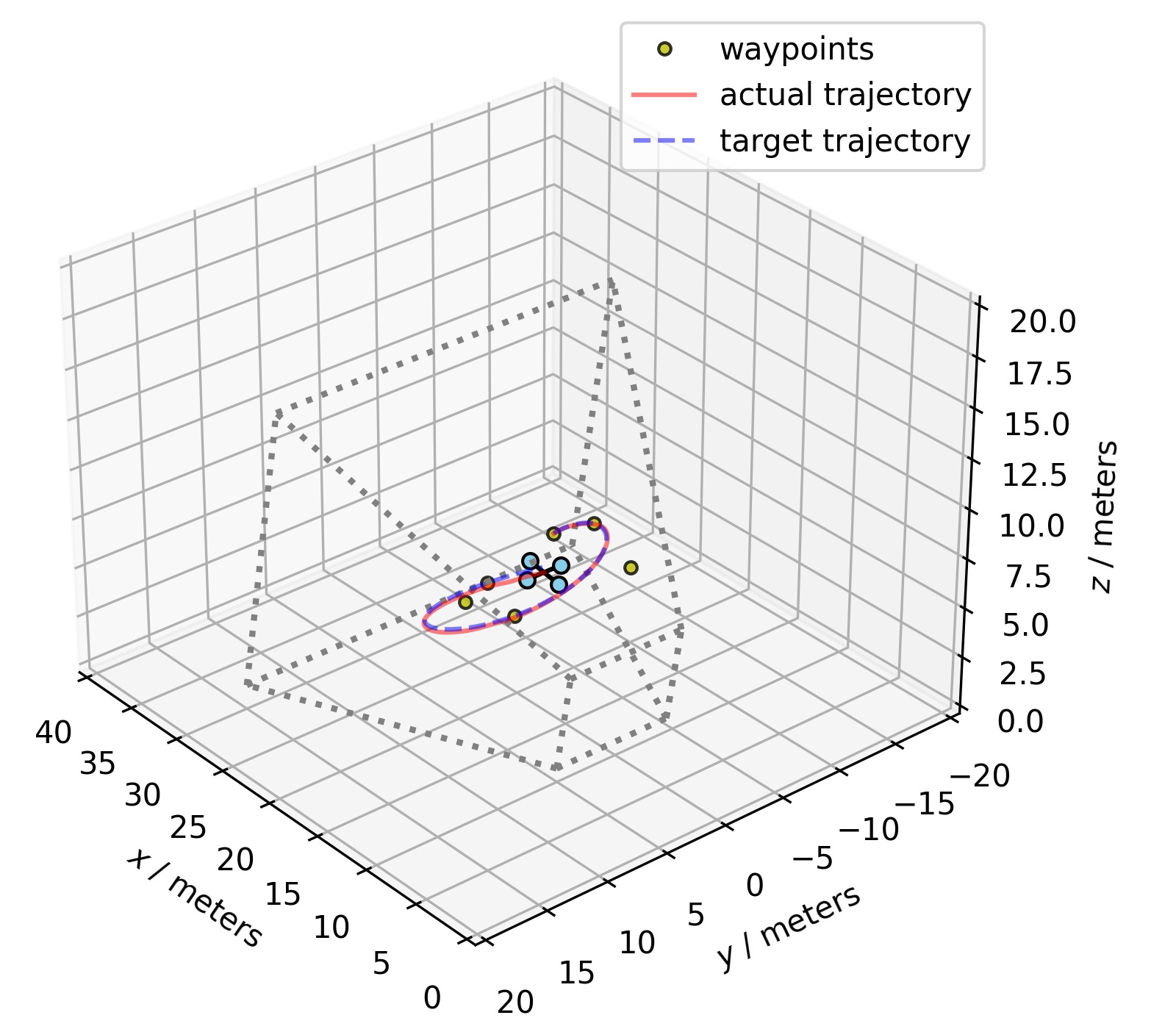}};
					\node (imag22) at (4,0) {	\includegraphics[width=0.4\textwidth]{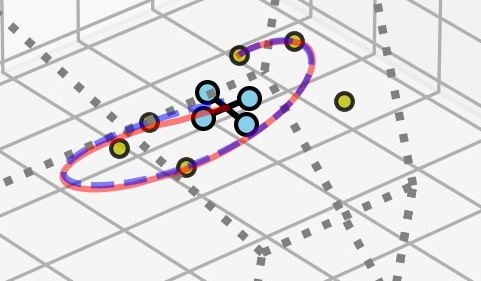}};
					\node[] (note1) at (4.0,2.1) {\small zoomed-in detail};	
					\draw[KITblue70,ultra thick,rounded corners, densely dotted] (-5.4,0.1) rectangle (-3.1,-1.2);
					\draw[KITblue70,ultra thick,rounded corners, densely dotted] (0.85,1.9) rectangle (7.2,-1.9);
				  \draw [ultra thick, KITblue70,densely dotted] (-3.1,-0.6) to [out=0, in=180] (0.85,0.0);					
					\end{scope}									
				\end{tikzpicture} 	
}	
\caption{Visualization of  a generated MST (target trajectory) along waypoints in the viewing frustum of a camera. The MST is highlighted in purple. The sampled waypoints in the viewing frustum of the camera are shown in yellow. The corresponding UAV flights with a PD controller as proposed from \cite{Michael_RAM_2010} are shown in red.}
\label{fig:ms_with_control} 	
\end{figure*}
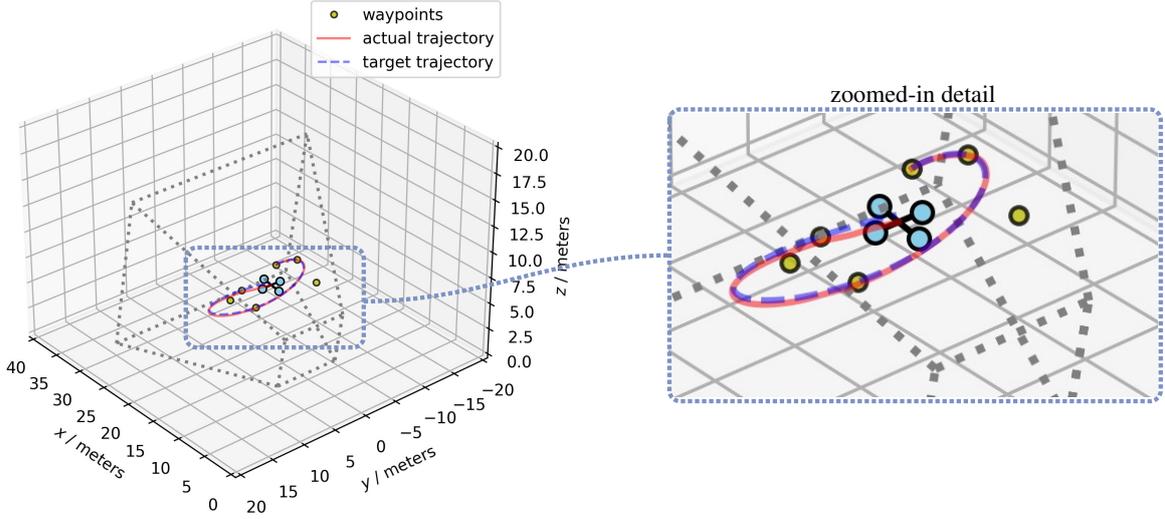

\section{\uppercase{Synthetic Data Generation}}
\label{sec:data_gen}
In this section the proposed approach for generating realistic, synthetic trajectory data is presented.

\textbf{Minimum Snap Trajectories (MST):} UAVs can not fly arbitrary trajectories due to the fact that they are dynamical systems with strict constraints on achievable velocities, accelerations and inputs. These constraints determine optimal trajectories with a series of waypoints in a set of positions and orientations in conjunction with control inputs \citep{Mellinger_phd_thesis_2020}. Thus, the goal of trajectory generation in controlling UAVs is to generate inputs to the motion control system, which ensures that the planned trajectory can be executed. Here, we apply an explicit optimization method that enables autonomous, aggressive, high-speed quadrotor flight through complex environments. In the remainder of this paper, the terms quadrotor and UAV are used interchangeably, although there exist various other UAV concepts. The principle procedure can be adapted to all other designs. Since we are only interested in the trajectory data, the actual control design can, to some extent, be neglected as long as the planned target trajectory is suitable for control.\\
Minimum snap trajectories (MST) have proven very effective as quadrotor trajectories since the motor commands and attitude accelerations of the UAV are proportional to the snap, or the fourth derivative, of the path \citep{Richter_ISRR_2016}. Of course, the difference between the target trajectory and the executed trajectory depends on the actual controller and physical limitations (e.g., maximum speed) of a UAV. Firstly, physical constraints can be considered in planning. Secondly, in most cases, a well-designed controller can closely follow the target trajectory. For our purpose, the trajectory of the actual flight can, in a way, be seen as only a slight variation of the target trajectory. However, quadrotor dynamics relying on four control inputs $(i_{1},\cdots,i_{4})$ for nested feedback control (inner attitude control loop and outer position control loop, see for example \cite{Michael_RAM_2010}) are differential flat \citep{Mellinger_ICRA_2011}. In other words, the states and the inputs can be written as functions of four selected \emph{flat outputs} and their derivatives. This facilitates the automated generation of trajectories since any smooth trajectory (with reasonably bounded derivatives) in the space of \emph{flat outputs} can be followed by the quadrotor. Following Mellinger et al., the \emph{flat outputs} are given by $p = [\vb{r}^\intercal, \psi ]^\intercal$, where $\vb{r} = [x, y, z]^\intercal$ are the coordinates of the center of mass in the world coordinate system and $\psi$ is the yaw angle. The \emph{flat outputs} at a given time $t$ are given by $p(t)$, which defines a smooth curve. A waypoint denotes a position in space along a yaw angle. Trajectory planning specifies navigating through $m$ waypoints at specified times by staying in a safe corridor. Trivial trajectories such as straight lines lead to discontinuities in higher-order motion. Thus, such trajectories are undesirable because they include infinite curvatures at waypoints, which require the quadrotor to stop at each waypoint. The differentiability of polynomial trajectories makes them a natural choice for use in a differentially flat representation of the quadrotor dynamics. Thus, for following a specific trajectory $p_{T}(t)=[\vb{r}_{T}^\intercal,\psi_{T}(t)]^\intercal$ (with controller for a UAV), the smooth trajectory $p_{T}(t)$ is defined as piecewise polynomial functions of order $n$ over $m$ time intervals:
\begin{equation}
\centering
p_{T}(t)=
    \left\{
    \begin{array}{@{}lr@{}}
      \sum^{n}_{i=0}p_{T_{i1}}(t^{i}) &  t_{0} \leqslant t \leqslant t_{1} \\
      \sum^{n}_{i=0}p_{T_{i2}}(t^{i})&t_{1} \leqslant t \leqslant t_{2}  \\
			\multicolumn{2}{c}{\vdots} \\
      \sum^{n}_{i=0}p_{T_{im}}(t^{i})& t_{m-1} \leqslant t \leqslant t_{m}
    \end{array}
    \right. 
\label{eq:piecewise_polynomial} 
\end{equation}

The goal is to find trajectories that minimize functionals which can be written using these basis functions. These kinds of problems can be solved with tools from the calculus of variations and are standard problems in robotics \citep{Craig_book}. Hence, in order to find the smooth target trajectory $p_{T}(t)_{tar}$,  the integral of the $k_{r}$\textsuperscript{th} derivative of position squared and the $k_{\psi}$\textsuperscript{th} derivative of yaw angle squared is minimized:
\begin{equation}
 p_{T}(t)_{tar} = \underset{p_{T}(t)}{\text{arg min}}  \int^{t_{m}}_{t_{o}} c_{r} \left\| \dv[k_{r}]{ \vb{r}_{T}}{t} \right\|^2 +  c_{\psi} \dv[k_{\psi}]{ \psi_{T}}{t}^2 \dd t
\label{eq:cost_mst} 
\end{equation}

Here, $c_{r}$ and $c_{\psi}$ are constants to make the integrand non-dimensional. Continuity of the $k_{r}$ derivatives of $\vb{r}_{T}$ and $k_{\psi}$ derivatives of $\psi_{T}$ is enforced as a criterion for smoothness. In other words, the continuity of the derivatives determines the boundary conditions at the waypoints. As mentioned above, some UAV control input depends on the fourth derivative of the positions and the second derivative of the yaw. Accordingly, $p_{T}(t)_{tar}$ is calculated by minimizing the integral of the square of the norm of the snap ($k_{r}=4$), and for the yaw angle, $k_{\psi}=2$ holds. 

The above problem can be formulated as a quadratic problem QP \citep{Bertsekas_book_1999}. Thereby, $p_{T_{ij}}= [ x_{T_{ij}}, y_{T_{ij}}, z_{T_{ij}}, \psi_{T_{ij}} ]^\intercal$ are written as a $4nm \times 1$ vector $\vec{c}$ with decision variables $\{x_{T_{ij}}, y_{T_{ij}}, z_{T_{ij}}, \psi_{T_{ij}}  \}$:
\begin{align}
\text{min  } \vec{c}^\intercal Q \vec{c} &+ \vec{f}^\intercal \vec{c} \nonumber \\
       \text{subject to  }  A\vec{c} &\leq \vec{b}  \text{.}  
\label{eq:QP_MST}
\end{align}
Here, the objective function incorporates the minimization of the functional while the constraints can be used to satisfy constraints on the flat outputs and their derivatives and thus constraints on the states and the inputs. 
The initial conditions, final condition, or intermediate condition on any derivative of the trajectory are specified as equality constraint in \ref{eq:QP_MST}. For a more detailed explanation on generating MSTs, how to incorporate corridor constraints, and how to calculate the angular velocities, angular accelerations, total thrust, and moments required over the entire trajectory for the controller, the reader is referred to the work of \cite{Mellinger_ICRA_2011}. \cite{Richter_ISRR_2016} presented an extended version of MST generation, where the solution of the quadratic problem is numerically more stable.\\
\textbf{Training Data Generation:} With the described method, we can generate MSTs suitable to aggressive maneuver flights for UAV control in a $3$D environment. In order to generate versatile trajectory data in image space, further steps are required. The overall generation pipeline is explained in the following. Firstly, a desired camera model with known intrinsic parameters is selected. The selection depends on the targeted sensor set-up of the detection and tracking system. In our case, we choose a pinhole camera model without distortion effects loosely orientated on the \emph{ANTI-UAV} dataset \citep{Jiang_arXiv_2021} with regard to an intermediate image resolution (in pixels) ($1176 \times 640 $) between the infrared ($640 \times 512$) and electro-optical camera ($1920 \times 1080$) resolutions of the \emph{ANTI-UAV} dataset. In case all camera parameters are known, the corresponding distortion coefficient should be considered. In the experiments the focal  length (in pixels) is set to $1240$, the principle point coordinates (in pixels) are set to $p_{x}=579$, $p_{y}= 212$. For setting up the external parameters, the camera is placed close to the ground plane uniformly sampled from $Uni(1m, 2m)$ to set the height above ground. The inclination angle is sampled from $Uni(\ang{10}, \ang{20})$. Given the fixed camera parameters, the viewing frustum is calculated for a chosen near distance to the camera center ($d_{near}=10m$) and a far distance to the camera center ($d_{far}=30m$).
For generating a single MST, a set of waypoints inside the viewing frustum is sampled. The number of waypoints is randomly varied between $3$ and $7$. The travel time for each segment is approximated by using the Euclidean distance between two waypoints and a sampled constant speed for the UAV $Uni(1m/s, 8m/s)$. The resulting straight-line trajectory serves as an initial solution of the MST calculation. The frame rate of the camera is sampled from $Uni(10fps, 20fps)$ for every run. In our experiments, we assumed a completely free space in the viewing frustum. As mentioned, corridor constraints, which can be used for simulating an object to fly through, can be integrated with the method of \cite{Mellinger_ICRA_2011}. For the synthetic training dataset, $1000$ MSTs are generated. The main steps for the synthetic data generation pipeline of a single run are as follows: 

\begin{itemize}
\item Selection of a desired camera model with known intrinsic parameters.    
\item Extrinsic parameters are sampled from the height and inclination angle distribution.
\item Calculation of the corresponding viewing frustum with $d_{near}$ and $d_{far}$.  
\item Sampling of waypoints inside viewing frustum. 
\item Estimation of the initial solution with fixed, sampled UAV speed.
\item MST trajectory generation using the method of \cite{Mellinger_ICRA_2011}.
\item Projection of the $3$D center of mass positions of the MST to image space using the camera parameters at fixed time intervals (reciprocal of the drawn frame rate of a single run).    
\end{itemize}

This procedure is repeated until the desired number of samples are generated. Note that sanity checks and abort criteria for the trajectory are applied at the end and during every run. For example, requirements on the minimum and maximum length of consecutive image points. In Figure \ref{fig:ms_with_control}, generated MSTs along sampled waypoints in a $3$D-environment are visualized. The viewing frustum for a fixed inclination angle of $\ang{15}$ is shown as a dotted gray line. The actual flight trajectory of a UAV is realized with a \emph{proportional–derivative} (PD) controller as is proposed by \cite{Michael_RAM_2010}. Although the controller design is relatively simple, the UAV can closely follow the target trajectory. Thus, directly using the planned MST seems to be a legitimate design choice. 
\begin{figure}[!ht]
\centering
\resizebox{0.9\columnwidth}{!}{	
				\begin{tikzpicture}[]
					\begin{scope}					
					\node (image1) at (0,0) {	\includegraphics[width=0.6\textwidth]{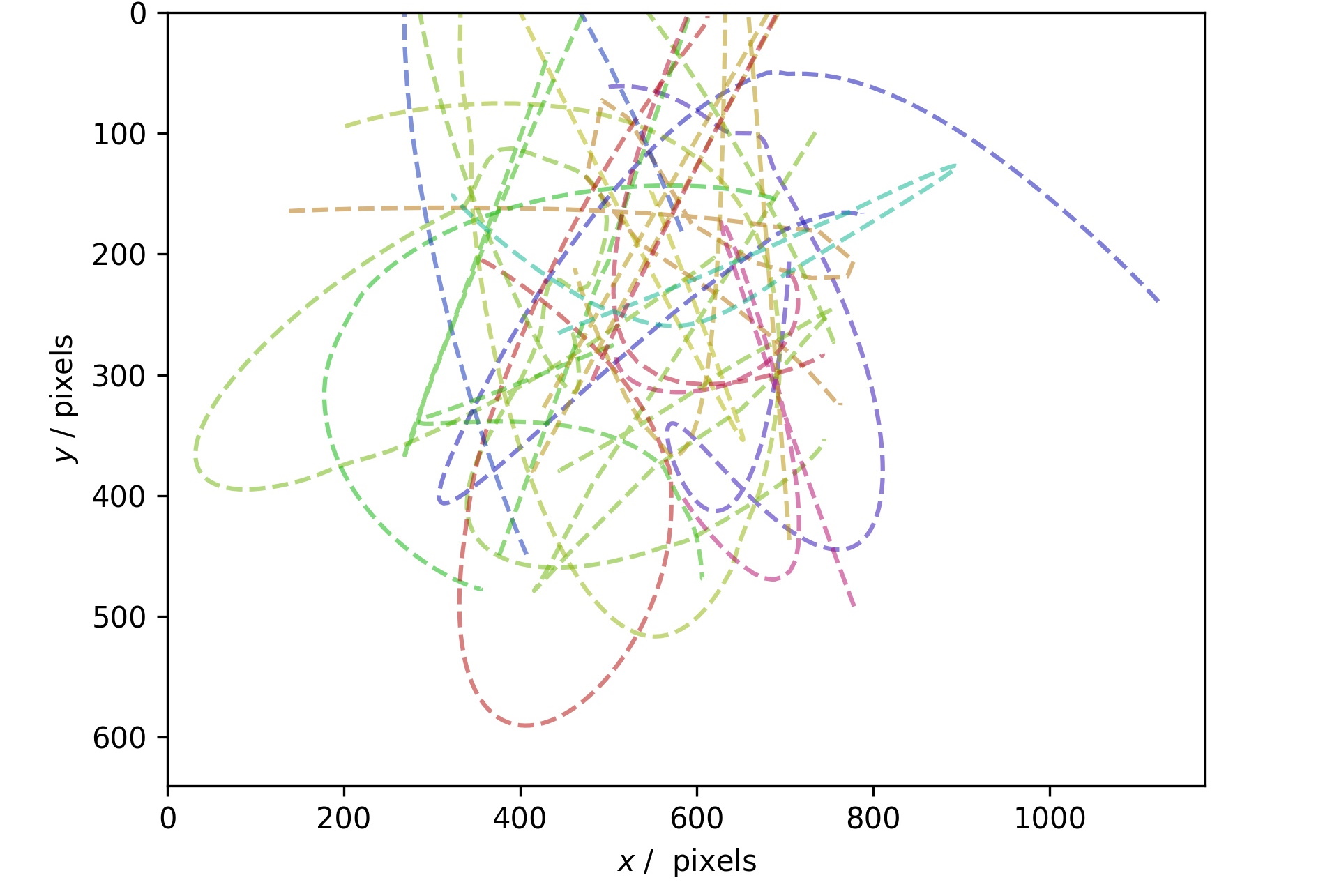}};						
					\end{scope}									
				\end{tikzpicture}
}	
\caption{Visualization of generated UAV trajectories as observed from the camera.}
\label{fig:synthetic_trajectory_samples} 	
\end{figure}

In Figure \ref{fig:synthetic_trajectory_samples}, exemplary generated training samples are depicted. The diversity and suitability of the generated synthetic data are analyzed in section \ref{sec:eval}. Before that, we will briefly introduce the deep learning-based reference prediction model.
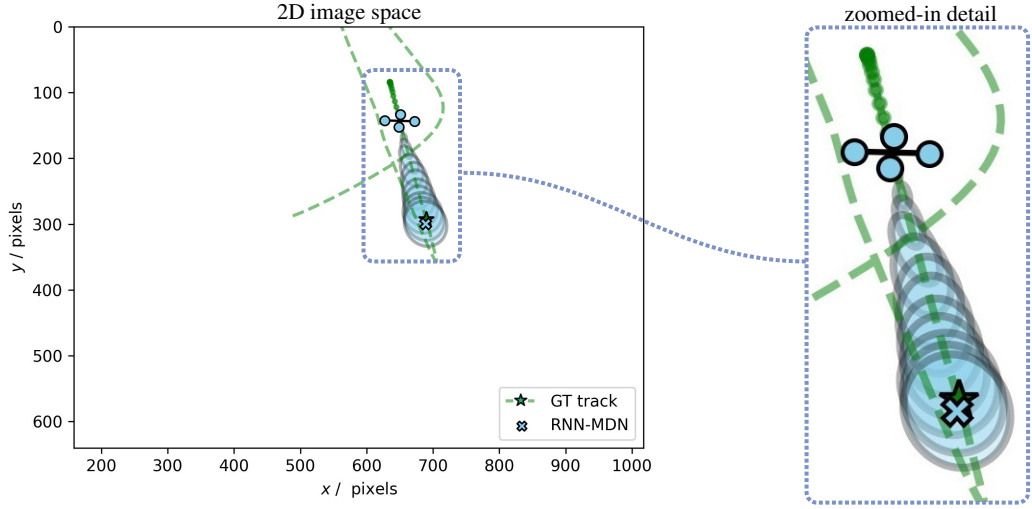
\begin{figure*}[!ht]
\centering
\resizebox{.9\textwidth}{!}{   		
				\begin{tikzpicture}[]
					\begin{scope}					
					\node (image1) at (-4,0) {	\includegraphics[width=0.65\textwidth]{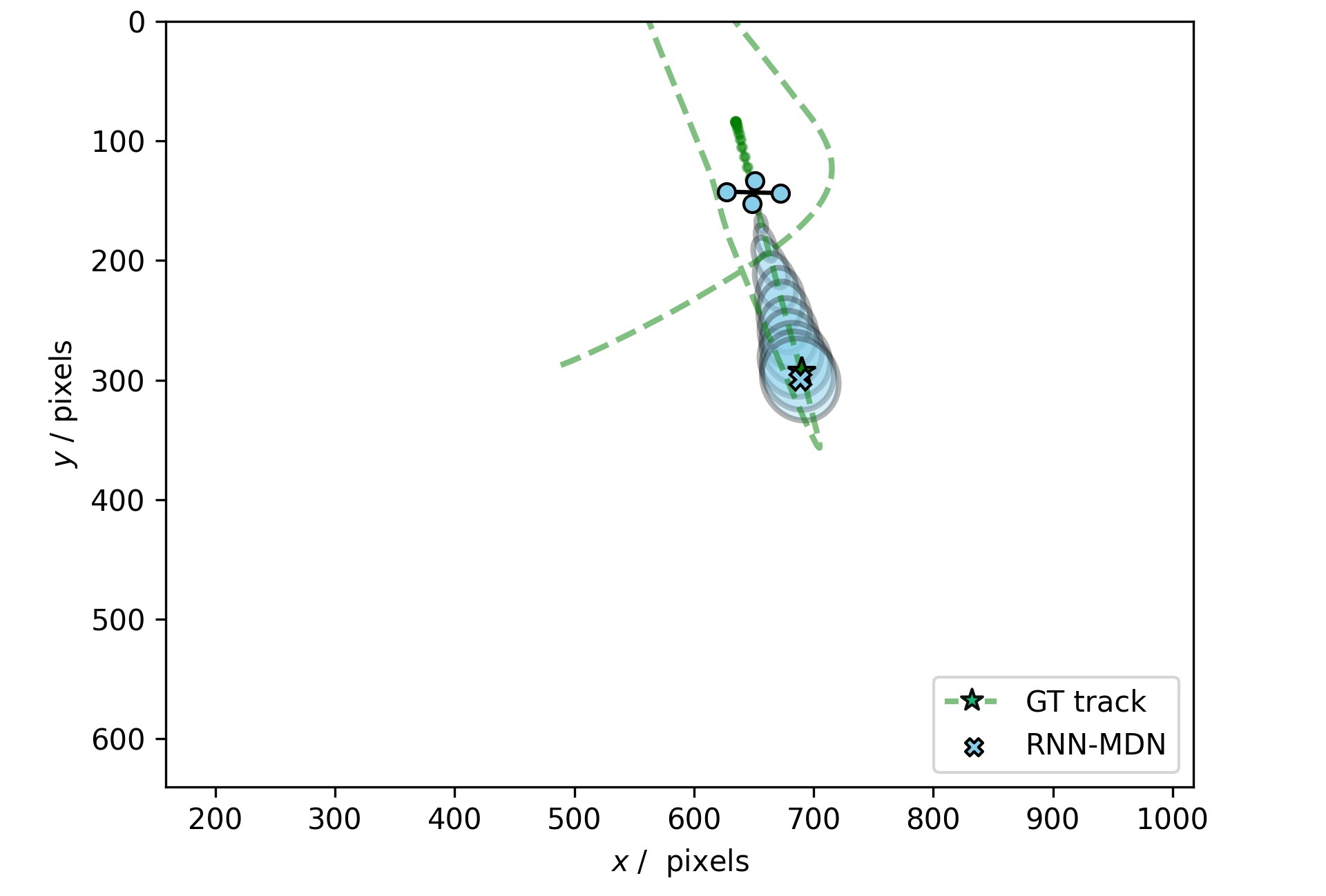}};
					\node (image2) at (4,-0.0) {	\includegraphics[width=0.2\textwidth]{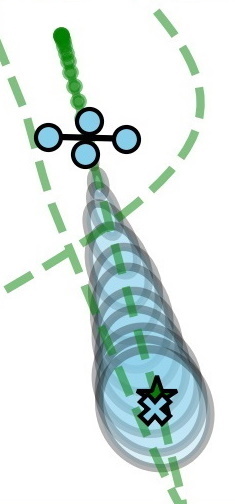}};				
					\node[] (note1) at (-4.0,3.5) {\small \text{$2$D  image space}  };				
					\node[] (note3) at (4.0, 3.5) { \small zoomed-in detail};	
					\draw[KITblue70,ultra thick,rounded corners, densely dotted] (-3.8,2.7) rectangle (-2.45,0.0);
					\draw[KITblue70,ultra thick,rounded corners, densely dotted] (2.4,3.3) rectangle (5.5,-3.4);
				  \draw [ultra thick, KITblue70,densely dotted] (-2.4,1.25) to [out=0, in=180] (2.4,0.0);					
					\end{scope}									
				\end{tikzpicture} 			
}	
\caption{Visualization of the predicted distribution of an RNN-MDN on a synthetic UAV trajectory in image space for $12$ steps into the future (on the left). A corresponding zoomed-in detail is shown on the right.}
\label{fig:rnn_mdn_synthetic} 	
\end{figure*}
\section{\uppercase{Prediction Model}}
\label{sec:model}
For trajectory prediction, an RNN-based model is considered, which predicts a distribution over the next $N$ image positions and is conditioned on the previous observations $\vec{u}^{0:t}$. $\vec{u}^{0:t}$ are the image coordinates from time $0$ up to $t$. As mentioned above, RNN-based models are a common choice from the variants of deep learning-based models for trajectory prediction. For generating the distribution over the next positions, the model outputs the parameter of a mixture density network (MDN). Such RNN-MDNs are applied to capture the motion of different object types. Originated from a model introduced by \cite{graves2014generating}, modified versions have been successfully utilized to predict pedestrian \citep{Alahi_CVPR_2016}, vehicle \citep{Deo_IV_2018} or cyclist \citep{Pool_TIV_2021} motions. Although these modified versions also incorporate some contextual cues (e.g., interactions from other objects), single objects motion is encoded with such an RNN-MDN variant.\\
Given an input sequence $\mathcal{U}$ of consecutive observed image positions $\vec{u}^t=(u^t,v^t)$ at time step $t$ along a trajectory, the model generates a probability distribution over future positions $\{ \vec{u}^{t+1}, \ldots, \vec{u}^{t+N}\}$. The model is realized as an RNN encoder. With an embedding of the inputs and using a single Gaussian component, the model can be defined as follows:
\begin{align}
\centering
 \vec{e}^{t}_{} = \text{EMB}&(\vec{u}^{t}; \vec{\Theta}_{e} ) \text{,} \nonumber \\
 \vec{h}^{t}_{} = \text{RNN}&(\vec{h}^{t-1}_{},\vec{e}^{t}_{}; \vec{\Theta}_{RNN} ) \text{,}\nonumber \\ 
 \{\hat{\vec{\mu}}^{t+n}, \hat{{\Sigma}}^{t+n}\}^{N}_{n=1} &= \text{MLP}(\vec{h}^{t}_{}; \vec{\Theta}_{MLP}) 
\label{eq:RNN_MDN} 
\end{align}

Here, $\text{RNN}(\cdot)$ is the recurrent network, $\vec{h}$ the hidden state of the RNN, $\text{MLP}(\cdot)$ the multilayer perceptron, and $\text{EMB}(\cdot)$ an embedding layer. $\vec{\Theta}$ represents the weights and biases of the MLP, EMB, or respectively of the RNN. The model is trained by maximizing the likelihood of the data given the output Gaussian distribution. This results in the following loss function:
\begin{align}
\mathcal{L}(\mathcal{U})_{} &=  - \sum_{n=1}^{N}  -\log \mathcal{N}( \vec{u}^{t+n} | \hat{\vec{\mu}}^{t+n}, \hat{\Sigma}^{t+n})  \text{.} 	 
\end{align} 

\textbf{Implementation Details:} The RNN-MDN is implemented using \emph{Tensorflow} \citep{tensorflow} and is trained for $2000$ epochs using an ADAM optimizer \citep{Kingma_ICLR_2015} with a decreasing learning rate, starting from $0.01$ with a learning rate decay of 0.95 and a decay factor of $\nicefrac{1}{10}$. The RNN hidden state and embedding dimension is $64$. For the experiments, the \emph{long short-term memory} (LSTM) \citep{Hochreiter_NC_1997} RNN variant is utilized.\\
An example prediction of the RNN-MDN on a synthetically generated UAV trajectory in image space is depicted in Figure \ref{fig:rnn_mdn_synthetic}. On the left, the predicted distribution for $12$ steps into the future is shown in blue. The corresponding ground truth position is marked as a green star. On the right, the corresponding  $3$D UAV trajectory is shown in green.
\section{\uppercase{Evaluation \& Analysis}}
\label{sec:eval}
This section analyzes the diversity of the generated synthetic data and the suitability for training deep-learning prediction models.
\begin{figure*}[t!]
	\setlength{\tabcolsep}{0.5pt}  
	\begin{center}	
		\begin{tabular}{cc|ccccc}
			\hline			
			\includegraphics[width=0.139\textwidth]{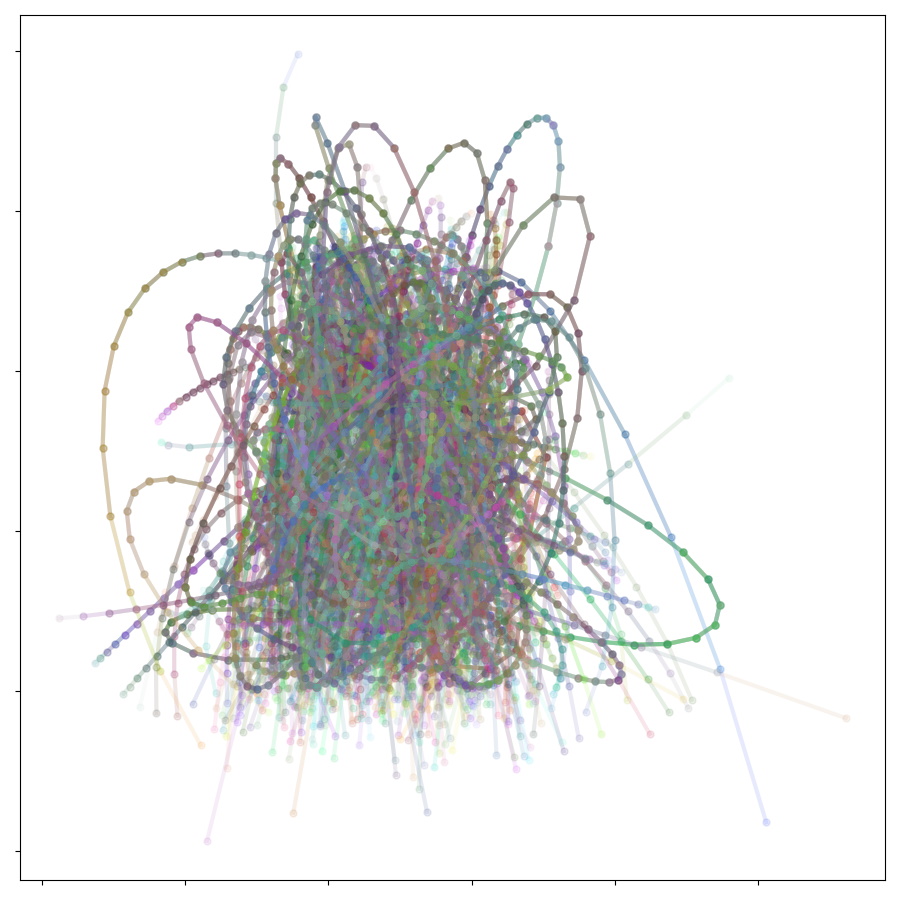} &
			\includegraphics[width=0.139\textwidth]{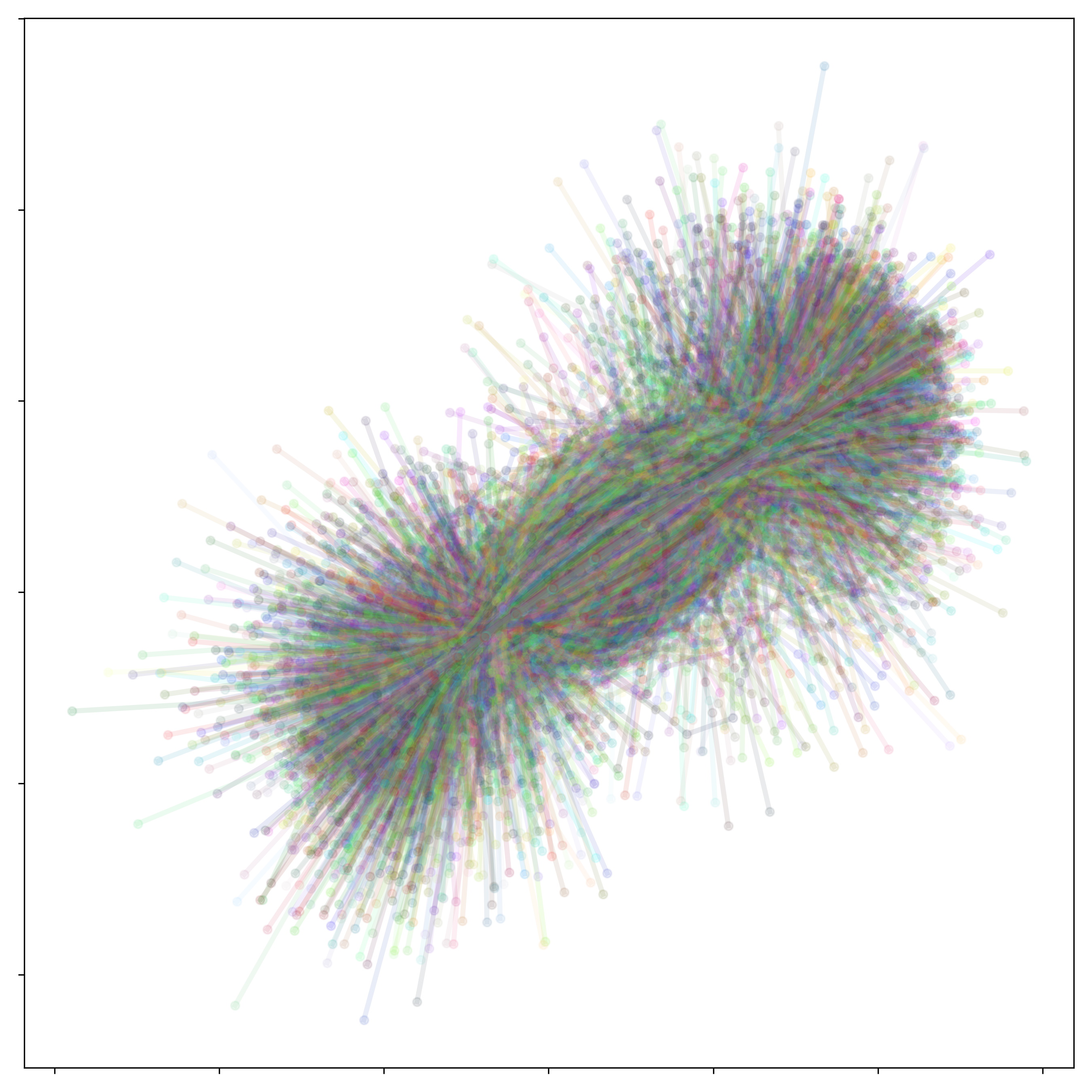} &	
			\includegraphics[width=0.139\textwidth]{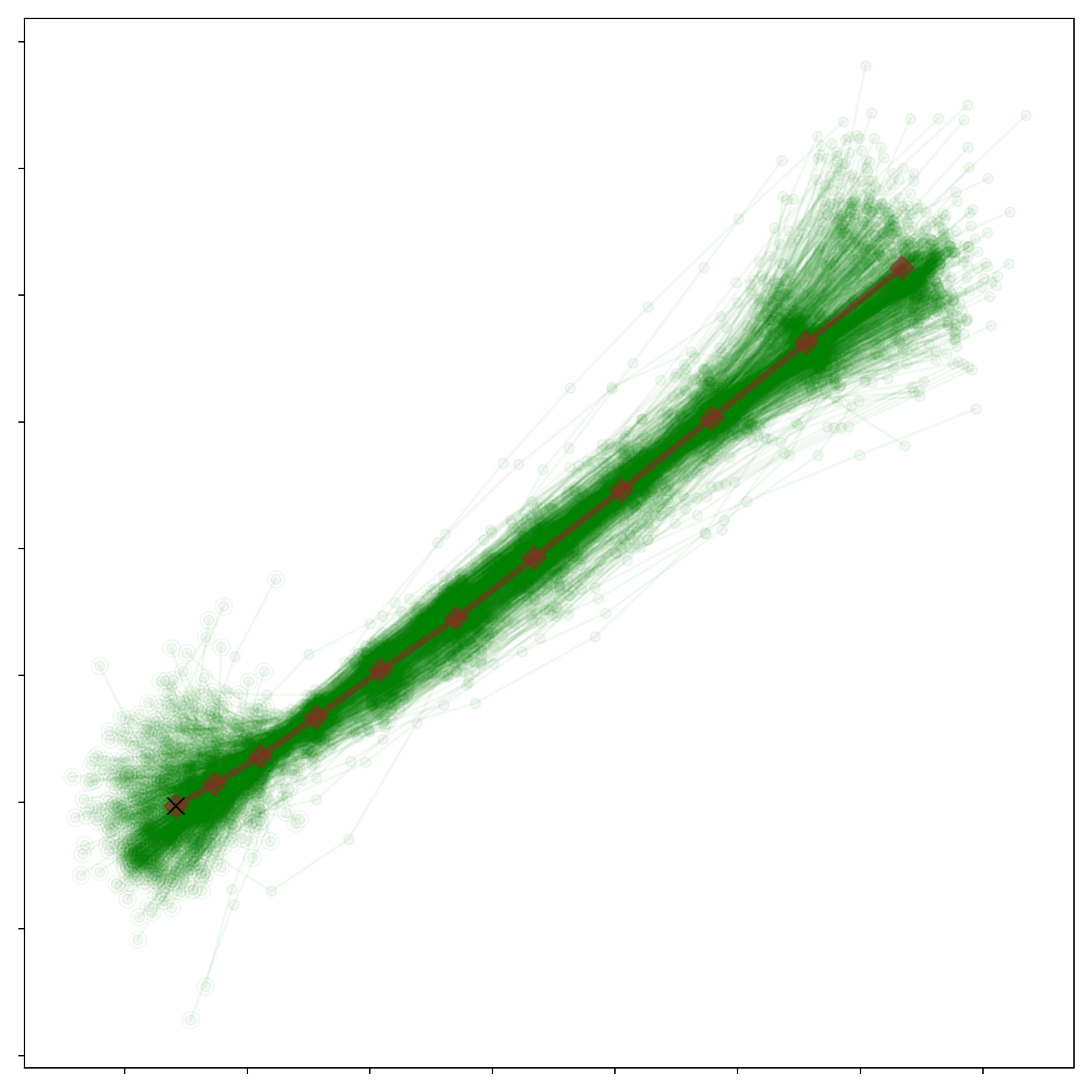} &
			\includegraphics[width=0.139\textwidth]{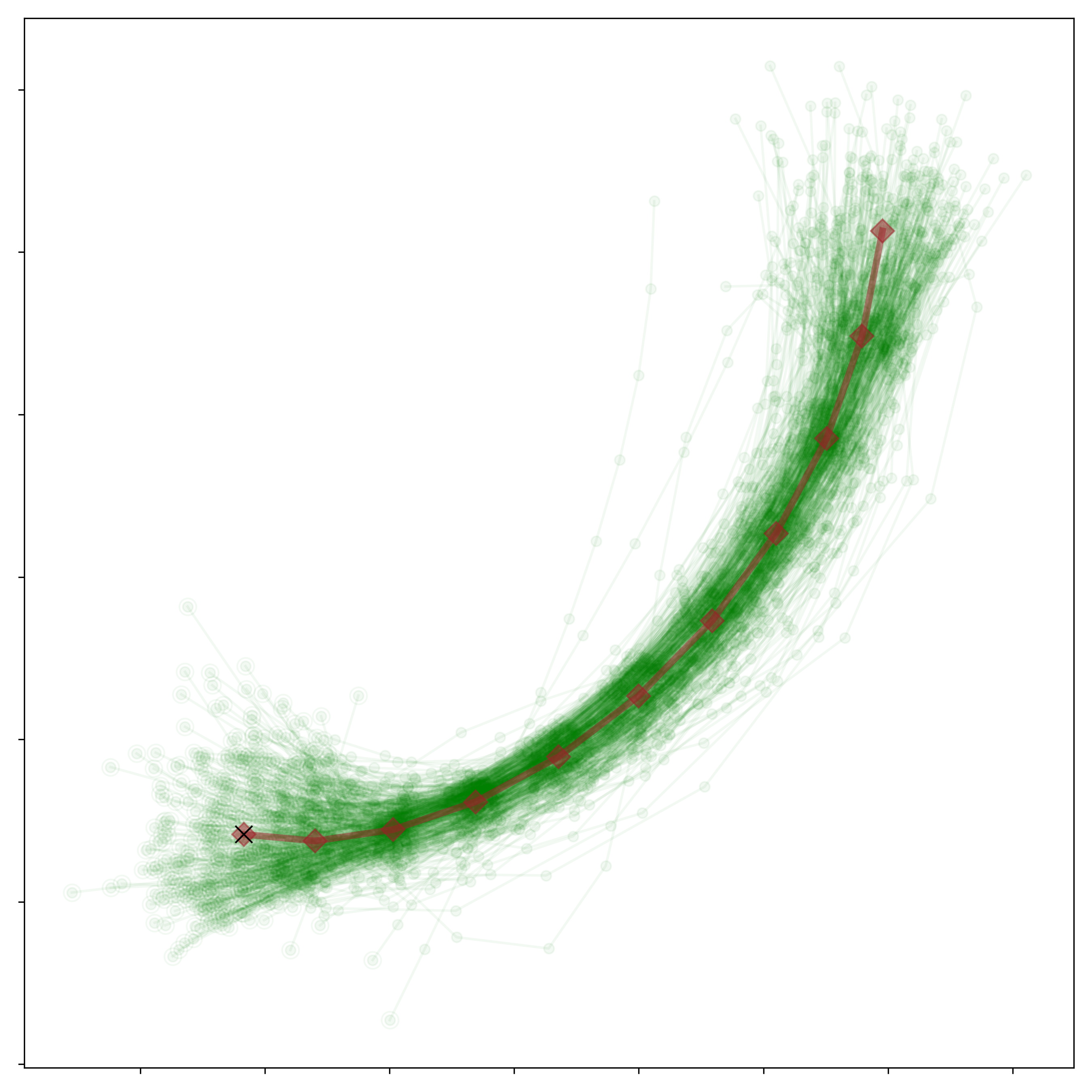} &
			\includegraphics[width=0.139\textwidth]{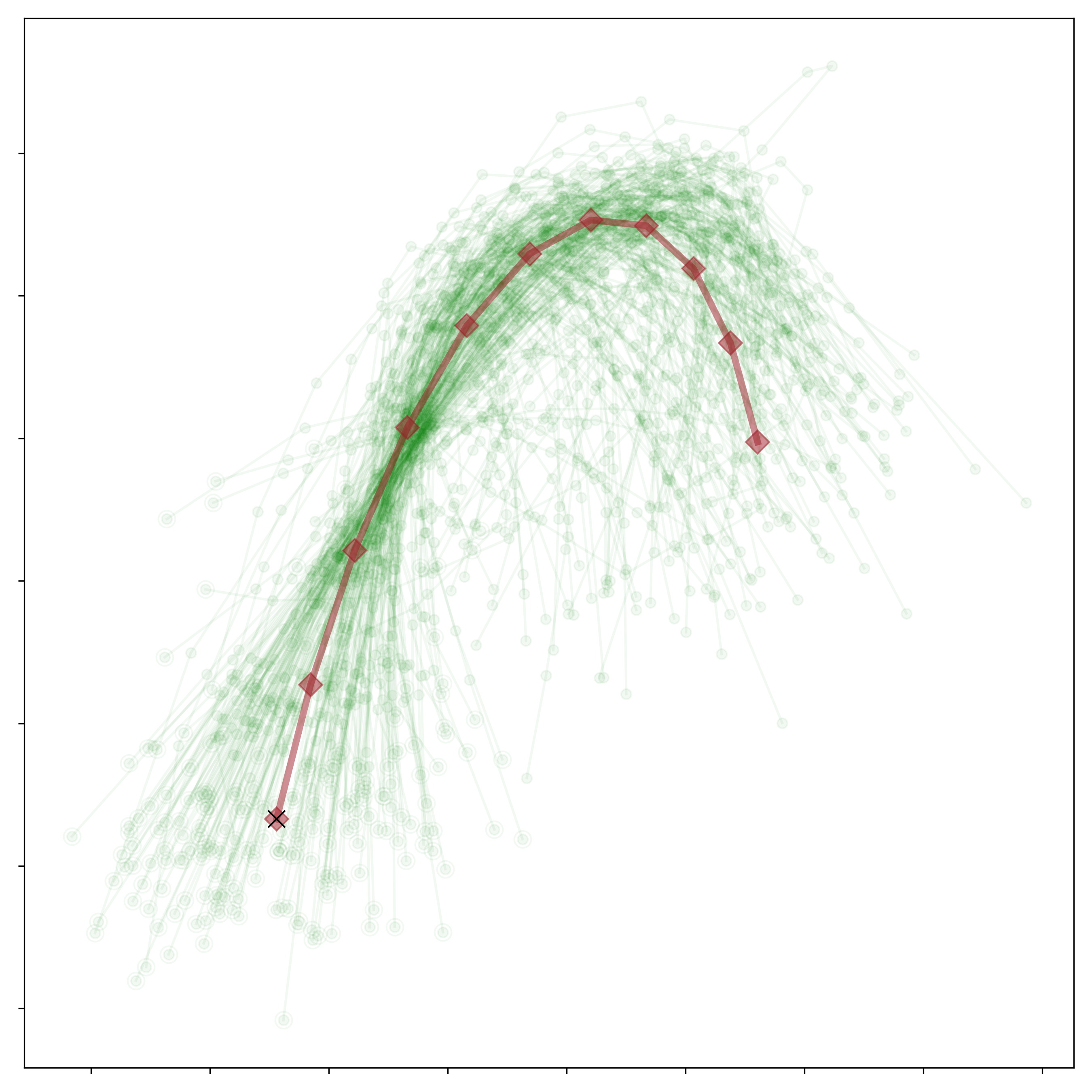} &
			\includegraphics[width=0.139\textwidth]{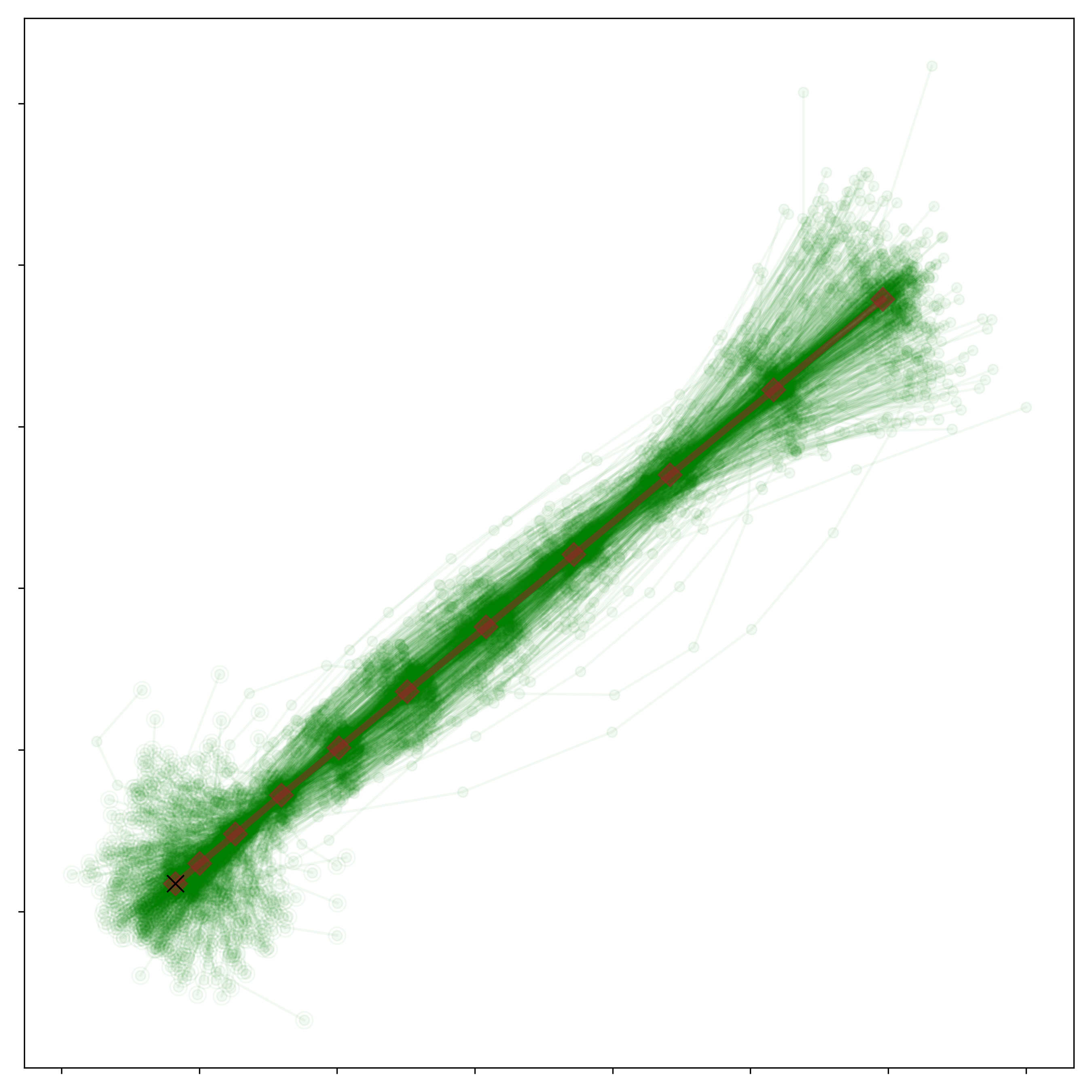} &
			\includegraphics[width=0.139\textwidth]{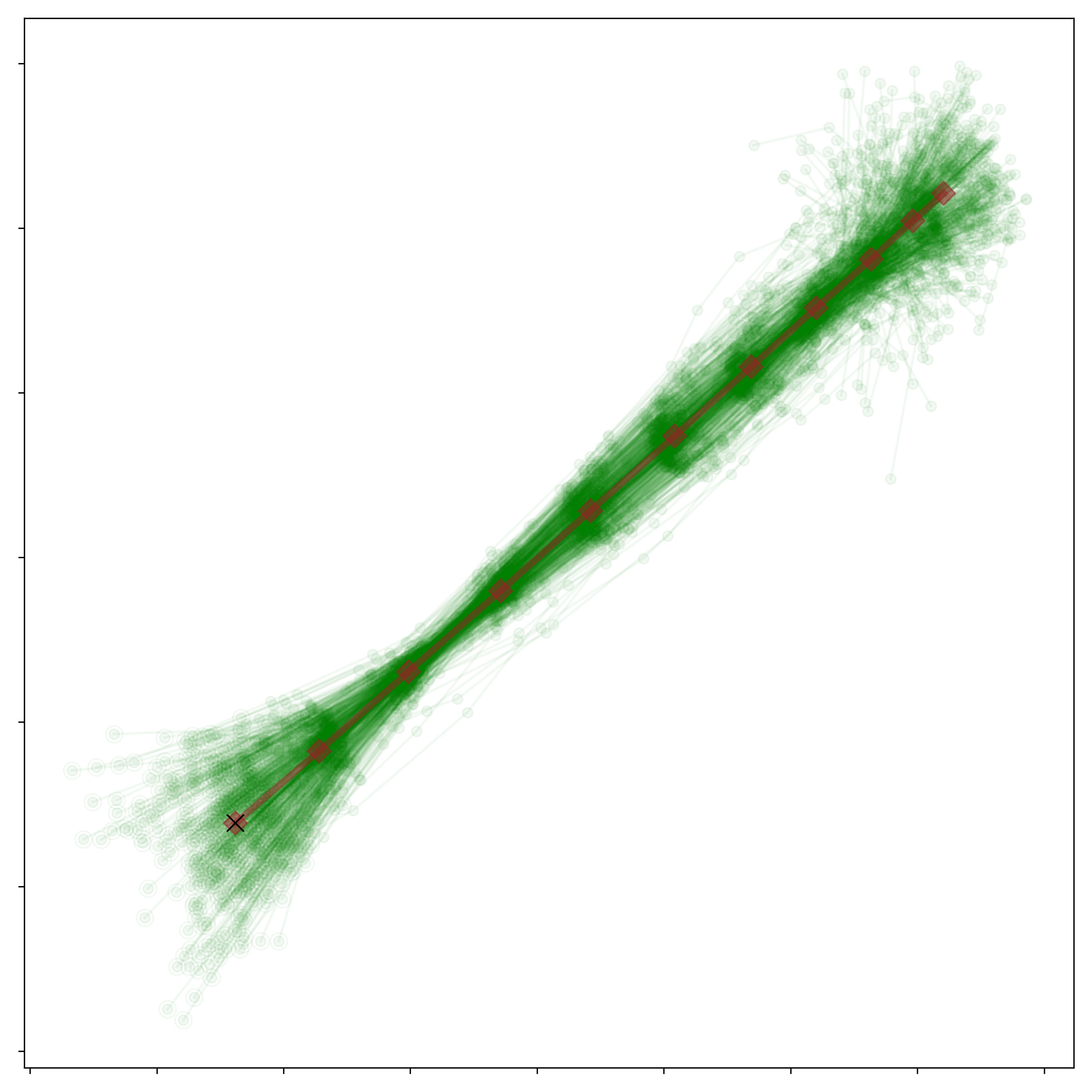}	\\
			\hline			
			\small Data & \small Aligned Data & \small \small Constant & \small Curve (l) & \small Curve (r) & \small Acceleration & \small Deceleration \\
			&& \multicolumn{5}{c}{\small Prototypes}
		\end{tabular}
	\end{center}
	\caption{Data, aligned data and learned prototypes for the synthetic UAV trajectories projected to image space. The prototypes represent different motion patterns (from left to right): Constant velocity, curvilinear motion (left and right curve), acceleration and deceleration.}
	\label{fig:results_alignment}
\end{figure*}
\begin{table*}[!h]
\resizebox{\textwidth}{!}{
	\begin{tabular}{| c | c c | c c | c c |}
      \hline
      \hline
			\multicolumn{1}{|c|}{} & \multicolumn{6}{c|}{ EO ($1920 \times 1080$)} \\
			 \hline
     	\multicolumn{1}{|c|}{} & \multicolumn{2}{c|}{ 8/8} &\multicolumn{2}{c|}{8/10}  & \multicolumn{2}{c|}{8/12}  \\
			\multicolumn{1}{|c|}{Approach} & {FDE/pixels}&$\sigma_{FDE}$/pixels & {FDE/pixels}&$\sigma_{FDE}$/pixels  & {FDE/pixels}&$\sigma_{FDE}$/pixels  \\
			\hline
			\rowcolor{gray!10}
      RNN-MDN               & 60.304 & 35.202 & 82.780  & 46.124 & 121.453 & 55.489     \\
			\rowcolor{KITblue!30}
      Kalman filter (CV)    & 81.061 & 60.333 & 110.041 & 84.530 & 179.459 &  104.408    \\
			\rowcolor{KITblue!30}
      Linear interpolation  & 86.998 & 67.113 & 119.106 & 89.067 & 183.558 & 108.522   \\
      \hline
			\hline
			 \hline
			\hline
			\multicolumn{1}{|c|}{} & \multicolumn{6}{c|}{IR ($640 \times 512$)} \\
			 \hline
			\multicolumn{1}{|c|}{} & \multicolumn{2}{c|}{ 8/8} &\multicolumn{2}{c|}{8/10}  & \multicolumn{2}{c|}{8/12}  \\
			\multicolumn{1}{|c|}{Approach} & {FDE/pixels}&$\sigma_{FDE}$/pixels & {FDE/pixels}&$\sigma_{FDE}$/pixels  & {FDE/pixels}&$\sigma_{FDE}$/pixels  \\
			\hline			
			\rowcolor{gray!10}
			RNN-MDN               & 20.849 & 18.604 & 41.378 & 21.100 & 61.459 & 24.490 \\
			\rowcolor{KITblue!30}
      Kalman filter (CV)    & 22.340 & 24.630 & 43.172 & 32.522 & 68.012 & 37.047 \\
			\rowcolor{KITblue!30}
      Linear interpolation  & 24.235 & 26.517 & 45.434 & 35.837 & 69.665 & 37.714 \\
      \hline
			\hline
	\end{tabular} 
	}
	\caption{Results for a comparison between the an RNN-MDN prediction model trained on the synthetic generated data, a Kalman filter with CV motion model, and using linear interpolation. The prediction is done for 8, 10, and 16 frames into the future.}
\label{result-table-anti_uav}
\end{table*}

\textbf{Diversity Analysis:} For analyzing the diversity of the synthetically generated data, we use the approach of \cite{Hug_IEEE_Access_2021}. The approach learns a representation of the provided trajectory data by first employing a spatial sequence alignment, which enables a subsequent learning vector quantization (LVQ) stage. Each trajectory dataset can be reduced to a small number of prototypical sub-sequences specifying distinct motion patterns, where each sample can be assumed to be a variation of these prototypes \citep{Hug_LHMP_2020}. Thus, the resulting quantized representation of the trajectory data, the prototypes, reflect basic motion patterns, such as constant or curvilinear motion, while variations occur primarily in position, orientation, curvature and scale. For further details on the dataset analysis methods, the reader is referred to the work of \cite{Hug_IEEE_Access_2021}. The resulting prototypes of the generated training data are depicted in Figure \ref{fig:results_alignment}.\\ 

The learned prototypes show that the generated synthetic data includes several different motion patterns. Besides, the diversity of the learned prototypes is visible. The dataset consists of, for example, distinguishable motion patterns reflecting constant velocity motion, curvilinear motion, acceleration, and deceleration.\\
\begin{figure*}[!th]
\centering
\resizebox{1.0\textwidth}{!}{
\begin{tabular}{cc}
       \multicolumn{1}{c}{\Large \text{IR Sequence: 20190925\_183946\_1\_6} } &
	    	\multicolumn{1}{c}{\Large \text{IR Sequence: 20190925\_152412\_1\_1} }\\					
				\begin{tikzpicture}[]
					\begin{scope}
					\node (image1) at (0,0) {	\includegraphics[width=0.8\textwidth]{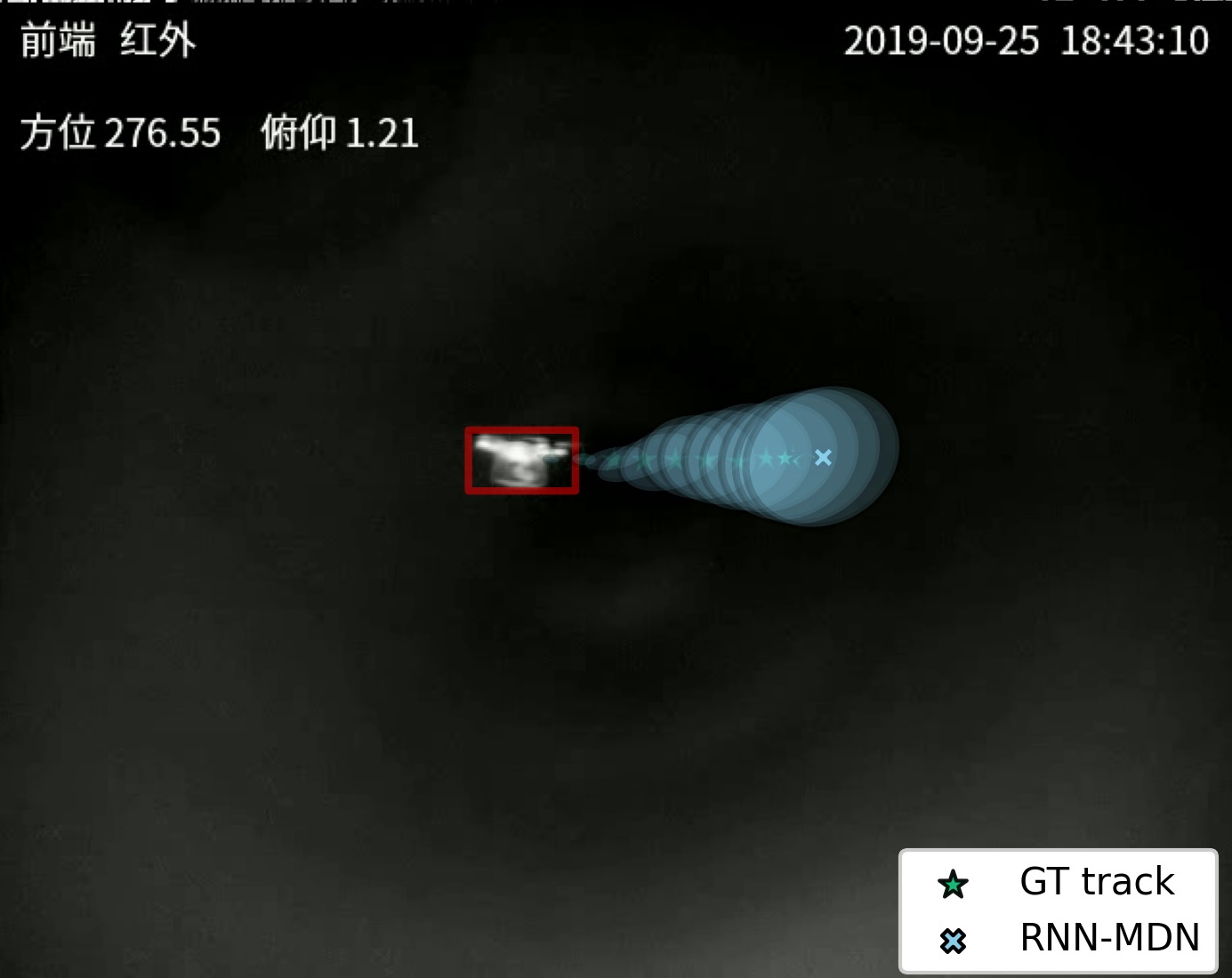}};					
					\end{scope}									
				\end{tikzpicture} &
				\begin{tikzpicture}
					\begin{scope}[font=\footnotesize]
					\node (image1) at (0,0) {	\includegraphics[width=0.8\textwidth]{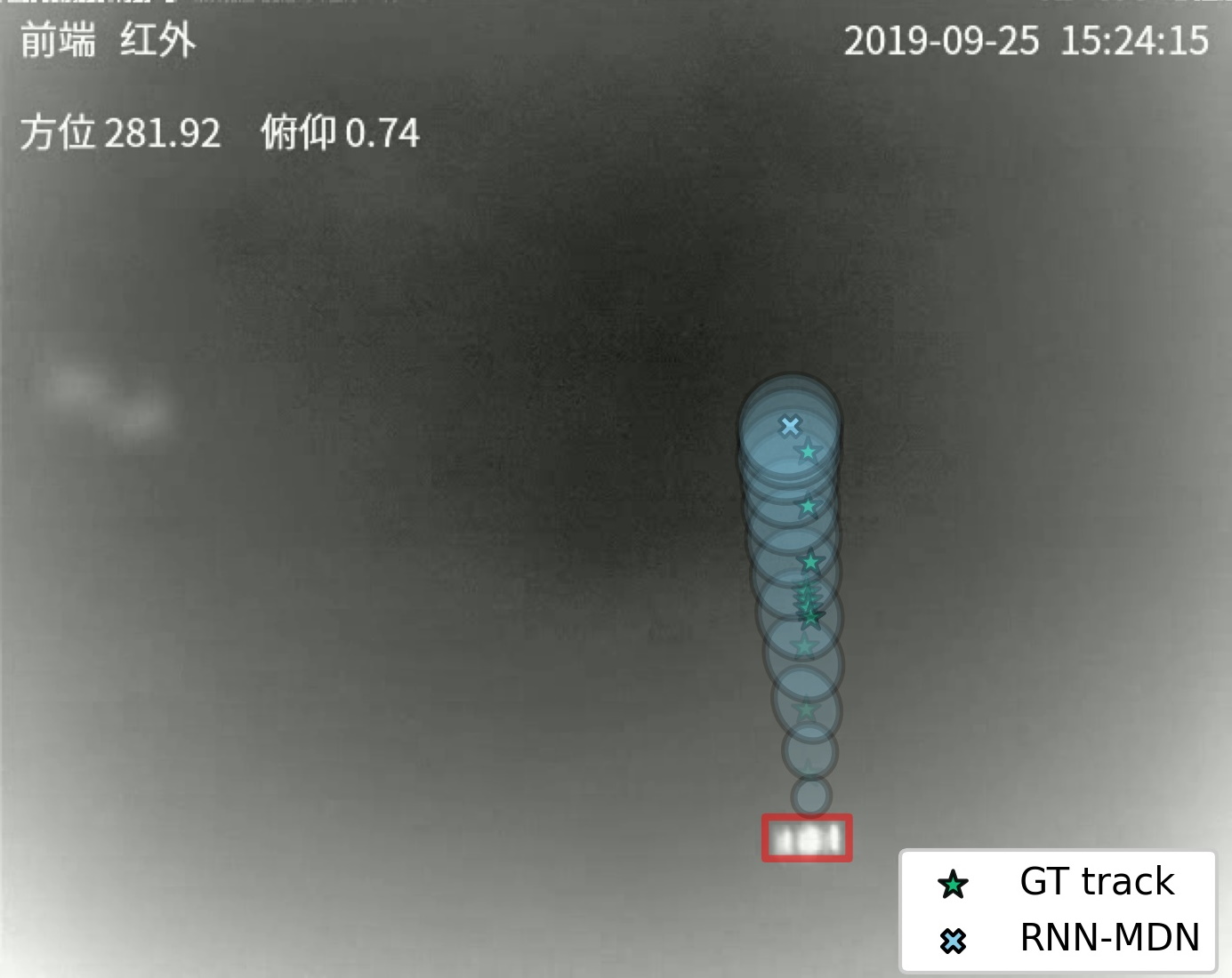}};						
					\end{scope}									
				\end{tikzpicture} \\				
				 \multicolumn{2}{c}{}\\
				 \multicolumn{1}{c}{\Large \text{EO Sequence: 20190925\_131530\_1\_5} } &
		\multicolumn{1}{c}{\Large \text{EO Sequence: 20190925\_133630\_1\_7} }\\				
				\begin{tikzpicture}[]
					\begin{scope}
					\node (image1) at (0,0) {	\includegraphics[width=0.9\textwidth]{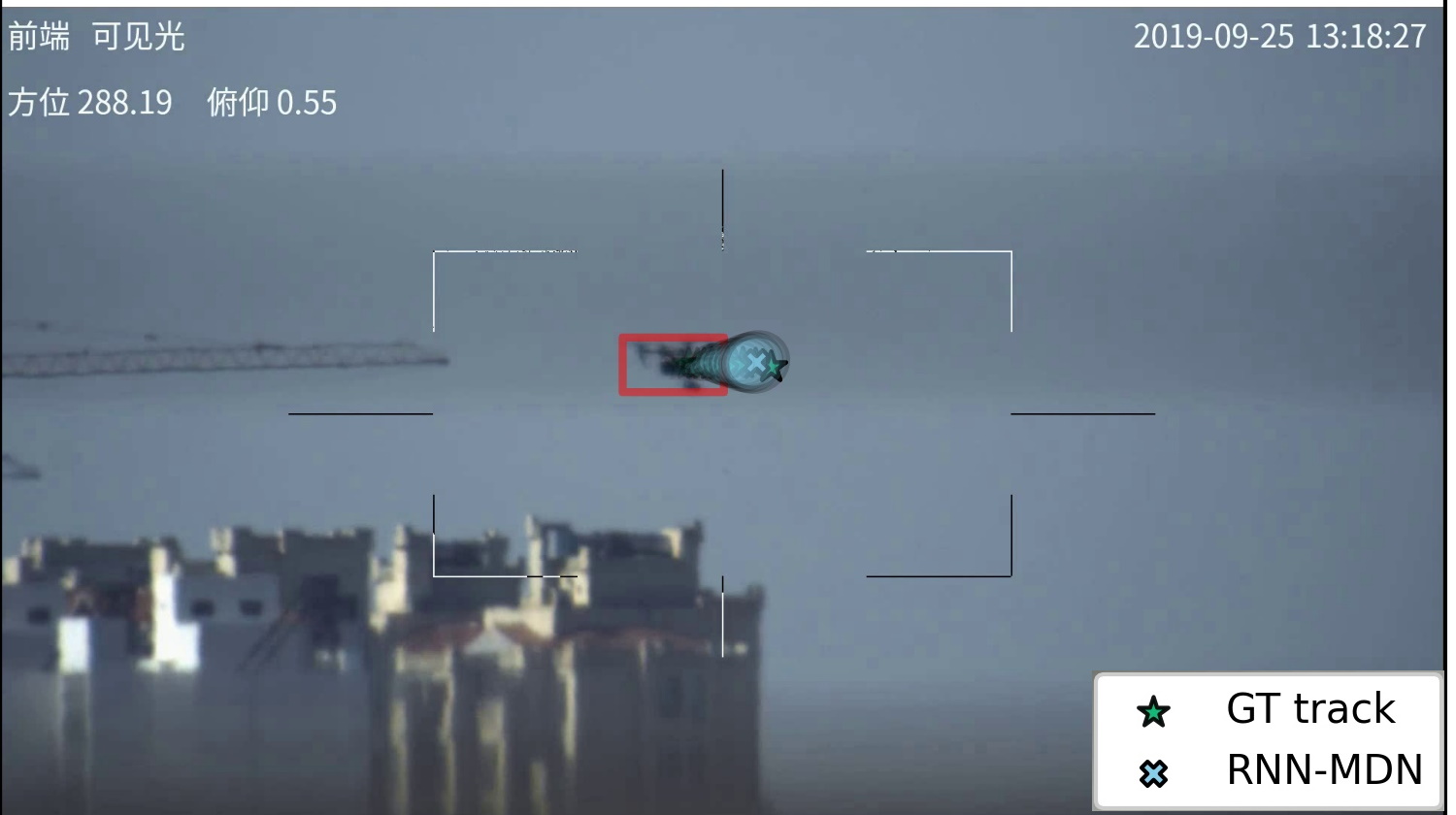}};						
					\end{scope}									
				\end{tikzpicture} &
				\begin{tikzpicture}
					\begin{scope}[font=\footnotesize]
					\node (image1) at (0,0) {	\includegraphics[width=0.9\textwidth]{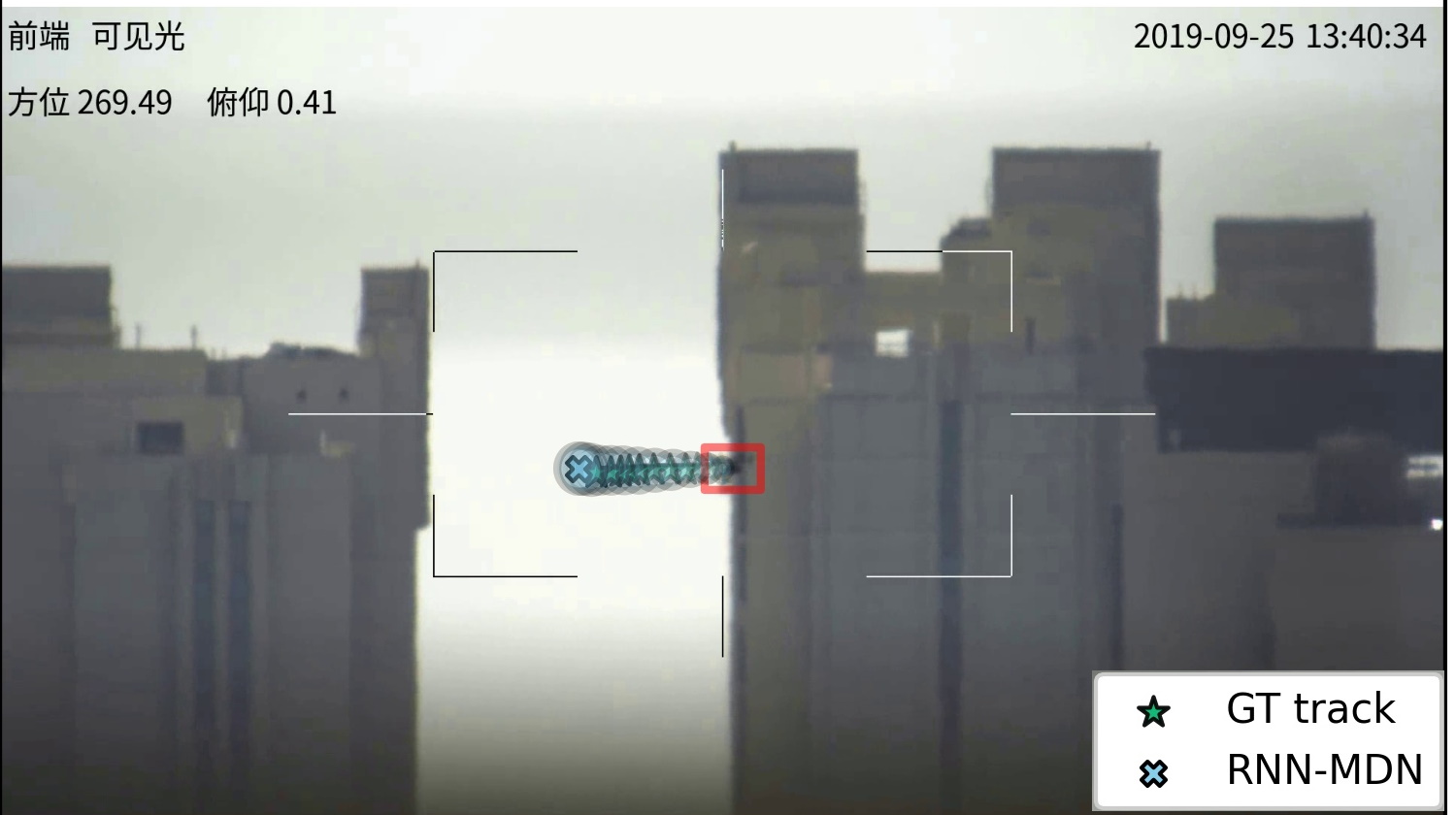}};							
					\end{scope}									
				\end{tikzpicture} \\	
\end{tabular}
}	
\caption{Example predictions for the \emph{Anti-UAV} dataset \cite{Jiang_arXiv_2021}. The top images show two samples from the IR sequences. The bottom images depict two samples from the EO sequences.}
\label{fig:examples_anti_uav} 	
\end{figure*}
\textbf{Suitability Analysis:} In order to analyze the suitability of the generated trajectory data, the RNN-MDN is trained as an exemplary deep learning-based prediction model according to section \ref{sec:model}. For evaluation, the real-world \emph{ANTI-UAV} dataset \citep{Jiang_arXiv_2021} is used. The dataset consists of $100$ high-quality, full HD video sequences (both electro-optical (EO) and infrared (IR)), spanning multiple occurrences of multi-scale UAVs, annotated with bounding boxes. As inputs for the prediction models, the center positions of the provided annotations are used. As classical reference models, a Kalman filter with a constant velocity (CV) motion model and linear interpolation are utilized. For the Kalman filter, the observation noise is assumed to be a white Gaussian noise process $\vec{w}^{t} \sim \mathcal{N}(0, (1.5pixels)^2)$. Thereby, the uncertainty in the annotation is considered. The process noise is modeled as the acceleration increment during a sampling interval (white noise acceleration model \cite{BarShalom_book_2002}) with $\sigma^{2}_{CV} = 0.5\nicefrac{pixels}{{frame}^2}$. For dealing with the minor annotation uncertainty, a small white Gaussian noise is added to the generated trajectory positions corresponding to the assumed observation noise. Since RNN are able to generalize from noisy inputs up to an extent (see for example \citep{Becker_ECCVW_2018}), the noisy trajectories are used for conditioning during training of the RNN-MDN. The performance is compared with the final displacement error (FDE) for three different time-horizons, in particular, $8$ frames, $10$ frames, and $12$ frames into the future. The FDE is calculated as the average Euclidean distance between the predicted final positions and the ground truth positions.\\
The results for the EO and the IR video sequences are summarized in table \ref{result-table-anti_uav}. Although the RNN-MDN is solely trained on the synthetically generated UAV data, the model could outperform the traditional reference models on the EO and the IR sequences of the \emph{ANTI-UAV} dataset. It should be noted that no camera motion compensation is applied. Thus, the results of all considered prediction models can be further improved. For longer prediction horizons, all errors are relatively high due to the maneuvering behavior of the UAVs. However, in case the prediction model is utilized to bridge detection failures for supporting the appearance model of the detection and tracking pipeline, the number of subsequent failures should be lower than the shown $12$ frames time horizon. Since the RNN-MDN relying on synthetic data achieved better performance than the reference models, it is better suited to anticipate the short-term UAV behavior. In Figure \ref{fig:examples_anti_uav}, the predicted distributions of future positions are visualized for the EO sequences or respectively IR sequences. The ground truth future positions are highlighted as green stars. The covariance ellipses around the predicted position are shown in blue. The final predicted positions are marked as a blue cross.
\vspace{-0.5cm}
\section{\uppercase{Conclusion}}
\label{sec:conclusion}
This paper presents an approach for generating synthetic trajectory data of UAVs in image space. By utilizing methods for planning maneuvering UAV flights, minimum snap trajectories along a sequence of $3$D waypoints are generated. By selecting the desired camera model with known camera parameters, the trajectories can be projected to the observer's perspective. To demonstrate the applicability of the synthetic trajectory data, we show that an RNN-MDN prediction model solely trained on the synthetically generated data is able to outperform classic reference models on a real-world UAV tracking dataset.

\bibliographystyle{apalike}
{\small
\bibliography{Becker_arxiv_DataGenUAV}}

\begin{thebibliography}{}

\bibitem[Abadi et~al., 2015]{tensorflow}
Abadi, M. et~al. (2015).
\newblock {TensorFlow}: Large-scale machine learning on heterogeneous systems.
\newblock Software available from tensorflow.org.

\bibitem[Alahi et~al., 2016]{Alahi_CVPR_2016}
Alahi, A., Goel, K., Ramanathan, V., Robicquet, A., Fei-Fei, L., and Savarese,
  S. (2016).
\newblock {S}ocial {LSTM}: {H}uman {T}rajectory {P}rediction in {C}rowded
  {S}paces.
\newblock In {\em Conference on Computer Vision and Pattern Recognition
  (CVPR)}, pages 961--971.

\bibitem[Amirian et~al., 2019]{Amirian_CVPR_W_2019}
Amirian, J., Hayet, J.-B., and Pettre, J. (2019).
\newblock {S}ocial {W}ays: {L}earning {M}ulti-{M}odal {D}istributions of
  {P}edestrian {T}rajectories {W}ith {GAN}s.
\newblock In {\em Conference on Computer Vision and Pattern Recognition (CVPR)
  Workshops}.

\bibitem[Bar-Shalom et~al., 2002]{BarShalom_book_2002}
Bar-Shalom, Y., Kirubarajan, T., and Li, X.-R. (2002).
\newblock {\em Estimation with Applications to Tracking and Navigation}.
\newblock John Wiley \& Sons, Inc., New York, NY, USA.

\bibitem[Becker et~al., 2018]{Becker_ECCVW_2018}
Becker, S., Hug, R., H\"{u}bner, W., and Arens, M. (2018).
\newblock Red: A simple but effective baseline predictor for the trajnet
  benchmark.
\newblock In {\em European Conference on Computer Vision (ECCV) Workshops}.
  Springer International Publishing.

\bibitem[Bertsekas, 1999]{Bertsekas_book_1999}
Bertsekas, D. (1999).
\newblock {\em Nonlinear Programming}.
\newblock Athena Scientific.

\bibitem[Bock et~al., 2019]{Bock_arXiv_2019}
Bock, J., Krajewski, R., Moers, T., Runde, S., Vater, L., and Eckstein, L.
  (2019).
\newblock The ind dataset: A drone dataset of naturalistic road user
  trajectories at german intersections.

\bibitem[Christnacher et~al., 2016]{Christnacher_SPIE_2016}
Christnacher, F., Hengy, S., Laurenzis, M., Matwyschuk, A., Naz, P., Schertzer,
  S., and Schmitt, G. (2016).
\newblock {Optical and acoustical UAV detection}.
\newblock In Kamerman, G. and Steinvall, O., editors, {\em Electro-Optical
  Remote Sensing X}, volume 9988, pages 83 -- 95. International Society for
  Optics and Photonics, SPIE.

\bibitem[Craig, 1989]{Craig_book}
Craig, J. (1989).
\newblock {\em Introduction to Robotics: Mechanics and Control}.
\newblock Addison-Wesley Longman Publishing Co., Inc., USA, 2nd edition.

\bibitem[{Deo} and {Trivedi}, 2018]{Deo_IV_2018}
{Deo}, N. and {Trivedi}, M.~M. (2018).
\newblock Multi-modal trajectory prediction of surrounding vehicles with
  maneuver based lstms.
\newblock In {\em 2018 IEEE Intelligent Vehicles Symposium (IV)}, pages
  1179--1184.

\bibitem[Giuliari et~al., 2020]{Giuliari_ICPR_2020}
Giuliari, F., Hasan, I., Cristani, M., and Galasso, F. (2020).
\newblock Transformer {N}etworks for {T}rajectory {F}orecasting.
\newblock In {\em International Conference on Pattern Recognition (ICPR)}.

\bibitem[Graves, 2014]{graves2014generating}
Graves, A. (2014).
\newblock Generating sequences with recurrent neural networks.

\bibitem[Gupta et~al., 2018]{Gupta_CVPR_2018}
Gupta, A., Johnson, J., Fei-Fei, L., Savarese, S., and Alahi, A. (2018).
\newblock {S}ocial {GAN}: {S}ocially {A}cceptable {T}rajectories with
  {G}enerative {A}dversarial {N}etworks.
\newblock In {\em Conference on Computer Vision and Pattern Recognition
  (CVPR)}. IEEE.

\bibitem[Hammer et~al., 2019]{Hammer_SPIE_2019}
Hammer, M., Borgmann, B., Hebel, M., and Arens, M. (2019).
\newblock {UAV detection, tracking, and classification by sensor fusion of a
  360° lidar system and an alignable classification sensor}.
\newblock In Turner, M.~D. and Kamerman, G.~W., editors, {\em Laser Radar
  Technology and Applications XXIV}, volume 11005, pages 99 -- 108.
  International Society for Optics and Photonics, SPIE.

\bibitem[Hammer et~al., 2018]{Hammer_SPIE_2018}
Hammer, M., Hebel, M., Laurenzis, M., and Arens, M. (2018).
\newblock {Lidar-based detection and tracking of small UAVs}.
\newblock In Buller, G.~S., Hollins, R.~C., Lamb, R.~A., and Mueller, M.,
  editors, {\em Emerging Imaging and Sensing Technologies for Security and
  Defence III; and Unmanned Sensors, Systems, and Countermeasures}, volume
  10799, pages 177 -- 185. International Society for Optics and Photonics,
  SPIE.

\bibitem[Hochreiter and Schmidhuber, 1997]{Hochreiter_NC_1997}
Hochreiter, S. and Schmidhuber, J. (1997).
\newblock {L}ong {S}hort-{T}erm {M}emory.
\newblock {\em Neural Computation}, 9(8):1735--1780.

\bibitem[Hug et~al., 2021]{Hug_IEEE_Access_2021}
Hug, R., Becker, S., Hübner, W., and Arens, M. (2021).
\newblock Quantifying the complexity of standard benchmarking datasets for
  long-term human trajectory prediction.
\newblock {\em IEEE Access}, 9:77693--77704.

\bibitem[Hug et~al., 2020]{Hug_LHMP_2020}
Hug, R., Becker, S., H\"{u}bner, W., and Arens, M. (2020).
\newblock A short note on analyzing sequence complexity in trajectory
  prediction benchmarks.
\newblock In {\em Workshop on Long-term Human Motion Prediction (LHMP)}.

\bibitem[{Jeon} et~al., 2017]{Jeon_EUSIPCO_2017}
{Jeon}, S., {Shin}, J., {Lee}, Y., {Kim}, W., {Kwon}, Y., and {Yang}, H.
  (2017).
\newblock Empirical study of drone sound detection in real-life environment
  with deep neural networks.
\newblock In {\em European Signal Processing Conference (EUSIPCO)}, pages
  1858--1862.

\bibitem[Jiang et~al., 2021]{Jiang_arXiv_2021}
Jiang, N., Wang, K., Peng, X., Yu, X., Wang, Q., Xing, J., Li, G., Zhao, J.,
  Guo, G., and Han, Z. (2021).
\newblock Anti-uav: A large multi-modal benchmark for uav tracking.

\bibitem[{Kartashov} et~al., 2020]{Kartashov_TCSET_2020}
{Kartashov}, V., {Oleynikov}, V., {Koryttsev}, I., {Sheiko}, S., {Zubkov}, O.,
  {Babkin}, S., and {Selieznov}, I. (2020).
\newblock Use of acoustic signature for detection, recognition and direction
  finding of small unmanned aerial vehicles.
\newblock In {\em 2020 IEEE 15th International Conference on Advanced Trends in
  Radioelectronics, Telecommunications and Computer Engineering (TCSET)}, pages
  1--4.

\bibitem[Kingma and Ba, 2015]{Kingma_ICLR_2015}
Kingma, D. and Ba, J. (2015).
\newblock Adam: A method for stochastic optimization.
\newblock In {\em International Conference on Learning Representations (ICLR)}.

\bibitem[Kothari et~al., 2020]{Kothari_arXiv_2020}
Kothari, P., Kreiss, S., and Alahi, A. (2020).
\newblock {H}uman {T}rajectory {F}orecasting in {C}rowds: {A} {D}eep {L}earning
  {P}erspective.
\newblock {\em arXiv preprint arXiv:2007.03639}.

\bibitem[Laurenzis et~al., 2020]{Laurenzis_SPIE_2020}
Laurenzis, M., Rebert, M., Schertzer, S., Bacher, E., and Christnacher, F.
  (2020).
\newblock {Prediction of MUAV flight behavior from active and passive imaging
  in complex environment}.
\newblock In {\em Laser Radar Technology and Applications}, volume 11410, pages
  10--17. International Society for Optics and Photonics, SPIE.

\bibitem[Mellinger, 2012]{Mellinger_phd_thesis_2020}
Mellinger, D. (2012).
\newblock {\em Trajectory Generation and Control for Quadrotors}.
\newblock PhD thesis, University of Pennsylvania.

\bibitem[{Mellinger} and {Kumar}, 2011]{Mellinger_ICRA_2011}
{Mellinger}, D. and {Kumar}, V. (2011).
\newblock Minimum snap trajectory generation and control for quadrotors.
\newblock In {\em International Conference on Robotics and Automation (ICRA)},
  pages 2520--2525.

\bibitem[{Michael} et~al., 2010]{Michael_RAM_2010}
{Michael}, N., {Mellinger}, D., {Lindsey}, Q., and {Kumar}, V. (2010).
\newblock The grasp multiple micro-uav testbed.
\newblock {\em IEEE Robotics Automation Magazine}, 17(3):56--65.

\bibitem[M{\"{u}}ller and Erdn{\"{u}}ß, 2019]{Mueller_SPIE_2019}
M{\"{u}}ller, T. and Erdn{\"{u}}ß, B. (2019).
\newblock {Robust drone detection with static VIS and SWIR cameras for day and
  night counter-UAV}.
\newblock In {\em Counterterrorism, Crime Fighting, Forensics, and Surveillance
  Technologies}, volume 11166, pages 58--72. International Society for Optics
  and Photonics, SPIE.

\bibitem[Nikhil and Morris, 2018]{Nikhil_ECCVW_2018}
Nikhil, N. and Morris, B. (2018).
\newblock Convolutional neural network for trajectory prediction.
\newblock In {\em The European Conference on Computer Vision (ECCV) Workshops}.

\bibitem[{Pool} et~al., 2021]{Pool_TIV_2021}
{Pool}, E., {Kooij}, J. F.~P., and {Gavrila}, D.~M. (2021).
\newblock Crafted vs. learned representations in predictive models - a case
  study on cyclist path prediction.
\newblock {\em IEEE Transactions on Intelligent Vehicles}, pages 1--1.

\bibitem[Rasouli, 2020]{Rasouli_arXiv_2020}
Rasouli, A. (2020).
\newblock {D}eep {L}earning for {V}ision-based {P}rediction: {A} {S}urvey.
\newblock {\em arXiv preprint arXiv:2007.00095}.

\bibitem[Richter et~al., 2016]{Richter_ISRR_2016}
Richter, C., Bry, A., and Roy, N. (2016).
\newblock {\em Polynomial Trajectory Planning for Aggressive Quadrotor Flight
  in Dense Indoor Environments}, pages 649--666.
\newblock Springer International Publishing, Cham.

\bibitem[{Rozantsev} et~al., 2017]{Rozantsev_TPAMI_2017}
{Rozantsev}, A., {Lepetit}, V., and {Fua}, P. (2017).
\newblock Detecting flying objects using a single moving camera.
\newblock {\em IEEE Transactions on Pattern Analysis and Machine Intelligence},
  39(5):879--892.

\bibitem[Rudenko et~al., 2020]{Rudenko_IJRR_2020}
Rudenko, A., Palmieri, L., Herman, M., Kitani, K.~M., Gavrila, D.~M., and
  Arras, K.~O. (2020).
\newblock Human {M}otion {T}rajectory {P}rediction: {A} {S}urvey.
\newblock {\em The International Journal of Robotics Research}.

\bibitem[Saleh, 2020]{Saleh_arXiv_2020}
Saleh, K. (2020).
\newblock {P}edestrian {T}rajectory {P}rediction using {C}ontext-{A}ugmented
  {T}ransformer {N}etworks.
\newblock {\em arXiv preprint arXiv:2012.01757}.

\bibitem[Samaras et~al., 2019]{Stamatios_MPDI_2019}
Samaras, S., Diamantidou, E., Ataloglou, D., Sakellariou, N., Vafeiadis, A.,
  Magoulianitis, V., Lalas, A., Dimou, A., Zarpalas, D., Votis, K., Daras, P.,
  and Tzovaras, D. (2019).
\newblock Deep learning on multi sensor data for counter uav applications—a
  systematic review.
\newblock {\em Sensors}, 19(22).

\bibitem[{Schumann} et~al., 2017]{Schumann_AVSS_2017}
{Schumann}, A., {Sommer}, L., {Klatte}, J., {Schuchert}, T., and {Beyerer}, J.
  (2017).
\newblock Deep cross-domain flying object classification for robust uav
  detection.
\newblock In {\em International Conference on Advanced Video and Signal Based
  Surveillance (AVSS)}, pages 1--6.

\bibitem[Schumann et~al., 2018]{Schumann_SPIE_2018}
Schumann, A., Sommer, L., Müller, T., and Voth, S. (2018).
\newblock {An image processing pipeline for long range UAV detection}.
\newblock In {\em Emerging Imaging and Sensing Technologies for Security and
  Defence III; and Unmanned Sensors, Systems, and Countermeasures}, volume
  10799, pages 186--194. International Society for Optics and Photonics, SPIE.

\bibitem[{Sommer} et~al., 2017]{Sommer_AVSS_2017}
{Sommer}, L., {Schumann}, A., {Müller}, T., {Schuchert}, T., and {Beyerer}, J.
  (2017).
\newblock Flying object detection for automatic uav recognition.
\newblock In {\em International Conference on Advanced Video and Signal Based
  Surveillance (AVSS)}, pages 1--6.

\bibitem[{Taha} and {Shoufan}, 2019]{Taha_IEEEAccess_2019}
{Taha}, B. and {Shoufan}, A. (2019).
\newblock Machine learning-based drone detection and classification:
  State-of-the-art in research.
\newblock {\em IEEE Access}, 7:138669--138682.

\bibitem[Unlu et~al., 2019]{Unlu_TCVA_2019}
Unlu, E., Zenou, E., Riviere, N., and Dupouy, P.-E. (2019).
\newblock Deep learning-based strategies for the detection and tracking of
  drones using several cameras.
\newblock {\em IPSJ Transactions on Computer Vision and Applications}, 11(1):7.

\bibitem[{Xiao} and {Zhang}, 2019]{Xiao_ICSAI_2019}
{Xiao}, Y. and {Zhang}, X. (2019).
\newblock Micro-uav detection and identification based on radio frequency
  signature.
\newblock In {\em International Conference on Systems and Informatics (ICSAI)},
  pages 1056--1062.

\end{thebibliography}

%

\end{document}